\documentclass[letterpaper, 10 pt, conference]{ieeeconf}  

\IEEEoverridecommandlockouts                              

\usepackage[colorlinks]{hyperref}       
\usepackage{url}            
\usepackage{graphicx}
\usepackage{multirow}
\usepackage{comment}
\usepackage{booktabs}
\usepackage{subcaption}
\usepackage{wrapfig}
\usepackage[table]{xcolor}

\title{\LARGE \bf
Simple-BEV: What Really Matters for Multi-Sensor BEV Perception?
}

\author{Adam W. Harley$^{1}$, Zhaoyuan Fang$^{2}$, Jie Li$^{3}$, Rares Ambrus$^{3}$, Katerina Fragkiadaki$^{2}$ \\ \url{https://simple-bev.github.io/}\vspace{-1em} %
\thanks{$^{1}$Stanford University, {\tt\small aharley@cs.stanford.edu}. $^{2}$Carnegie Mellon University, {\tt\small \{zhaoyuaf, katef\}@cs.cmu.edu}
{$^{3}$Toyota Research Institute, {\tt\small \{jie.li, rares.ambrus\}@tri.global}}}
}

\begin{document}

\newcommand\adam[1]{\textcolor{magenta}{#1}}
\newcommand\todo[1]{\textcolor{red}{#1}}
\newcommand\red[1]{\textcolor{red}{#1}}
\newcommand\gist[1]{\textcolor{cyan}{#1}}
\newcommand\cyan[1]{\textcolor{cyan}{#1}}
\newcommand\gray[1]{\textcolor{gray}{#1}}
\newcommand\teal[1]{\textcolor{teal}{#1}}
\newcommand\pink[1]{\textcolor{magenta}{#1}}
\newcommand\katef[1]{\textcolor{magenta}{katef:#1}}

\newcommand{\model}{\text{DPV}}

\newcommand{\M}{\mathcal{M}} 
\newcommand{\R}{\mathcal{R}} 
\newcommand{\Mt}{\mathcal{M}^{(t)}} 
\newcommand{\Mz}{\mathcal{M}^{(0)}} 
\newcommand{\Mo}{\mathcal{M}^{(1)}} 
\newcommand{\Mi}{\mathcal{M}^{(i)}} 
\newcommand{\Mj}{\mathcal{M}^{(j)}} 
\newcommand{\Mto}{\mathcal{M}^{(t+1)}} 

\newcommand{\Tt}{\mathcal{T}^{(t)}} 
\newcommand{\Tz}{\mathcal{T}^{(0)}} 
\newcommand{\To}{\mathcal{T}^{(1)}} 

\newcommand{\I}{I} 
\newcommand{\D}{D} 

\newcommand{\loss}{\mathcal{L}}

\newcommand{\Ot}{{O}^{(t)}}
\newcommand{\Oz}{{O}^{(1)}}
\newcommand{\Ct}{{O}^{(t)}}
\newcommand{\Cht}{\hat{{O}}^{(t)}}
\newcommand{\Iht}{\hat{{I}}^{(t)}}

\renewcommand{\arraystretch}{1.2}

\maketitle
\thispagestyle{empty}
\pagestyle{empty}

\begin{abstract}
\looseness=-1 
Building 3D perception systems for autonomous vehicles that do not rely on high-density LiDAR is a critical research problem because of the expense of LiDAR systems compared to cameras and other sensors. Recent research has developed a variety of camera-only methods, where features are differentiably ``lifted'' from the multi-camera images onto the 2D ground plane, yielding a ``bird's eye view'' (BEV) feature representation of the 3D space around the vehicle. This line of work has produced a variety of novel ``lifting'' methods, but we observe that other details in the training setups have shifted at the same time, making it unclear \textit{what really matters} in top-performing methods. We also observe that using cameras alone is not a real-world constraint, considering that additional sensors like \textit{radar} have been integrated into real vehicles for years already. In this paper, we first of all attempt to elucidate the high-impact factors in the design and training protocol of BEV perception models. We find that batch size and input resolution greatly affect performance, while lifting strategies have a more modest effect---even a simple parameter-free lifter works well. Second, we demonstrate that radar data can provide a substantial boost to performance, helping to close the gap between camera-only and LiDAR-enabled systems. We analyze the radar usage details that lead to good performance, and invite the community to re-consider this commonly-neglected part of the sensor platform. 

\end{abstract}

\section{Introduction}

\looseness=-1 There is great interest in building 3D-aware perception systems for \textit{LiDAR-free} autonomous vehicles, whose sensor platforms typically constitute multiple RGB cameras, and multiple radar units. 
While LiDAR data enables highly accurate 3D object detection~\cite{yin2021center, yin2021multimodal}, 
the sensors themselves are arguably too expensive for large-scale deployment~\cite{li2022exploiting}, especially as compared to current camera and radar units. 
Most current works focus on producing an accurate ``bird's eye view'' (BEV) semantic representation of the 3D space surrounding the vehicle, using multi-view camera input alone. This representation  captures the information required for driving-related tasks, such as navigation, obstacle detection, and moving-obstacle forecasting. We have seen extremely rapid progress in this domain: 
for example, BEV vehicle semantic segmentation IOU improved from 23.9~\cite{pan2020cross} to 44.4~\cite{li2022bevformer} in just two years!

\looseness=-1 While this progress is encouraging, the focus on innovation and accuracy has come at the cost of system simplicity, and risks obscuring ``what really matters'' for performance~\cite{bewley2016simple, weng20203d}.  
There has been a particular focus on innovating new techniques for ``lifting'' features from the 2D image plane(s) onto the BEV plane. For example, some work has explored 
using homographies to warp features directly onto the ground plane \cite{can2022understanding}, 
using depth estimates to place features at their approximate 3D locations \cite{schulter2018learning,philion2020lift}, 
using MLPs with various geometric biases \cite{hendy2020fishing,saha2021enabling,gosala2022bird}, and most recently, using geometry-aware transformers~\cite{saha2021translating} and deformable attention across space and time~\cite{li2022bevformer}. {At the same time, implementation details have gradually shifted toward using more-powerful backbones and higher-resolution inputs, making it difficult to measure the actual impact of these developments in lifting.} 
We propose a model where the ``lifting'' step is \textbf{parameter-free} and does not rely on depth estimation: we simply define a 3D volume of coordinates over the BEV plane, project these coordinates into all images, and average the features sampled from the projected locations. When this simple model is tuned well, it exceeds the performance of state-of-the-art models while also being faster and more parameter-efficient. 
We measure the independent effects of batch size, image resolution, augmentations, and 2D-to-BEV lifting strategy, and show empirically that good selection of input resolution and batch size can improve performance by more than 10 points (all other factors held equal), while the difference between the worst and best lifting methods is only 4 points -- which is particularly surprising because lifting methods have been the main focus of earlier work. 

\looseness=-1 We further show that results can be substantially improved by incorporating input from \textbf{radar}. While recent efforts have focused on using cameras and/or LiDAR, we note that radar sensors have been integrated into real vehicles for years already~\cite{PARKER2017253}, and \textit{cameras plus radar} performs far better than cameras alone.  
While using cameras alone may give the task a certain purity (requiring metric 3D estimates from 2D input), it does not reflect the reality of autonomous driving, where noisy metric data is freely available, not only from radar but from GPS and odometry. 
The few recent works that discuss radar in the context of semantic BEV mapping have concluded that the data is often too sparse to be useful \cite{li2022exploiting, hendy2020fishing}. 
We identify that these prior works evaluated the use of \textit{radar alone}, avoiding the multi-modal fusion problem, and perhaps missing the opportunity for RGB and radar to complement one another. 
We introduce a simple RGB+radar fusion strategy (rasterizing the radar in BEV and concatenating it to the RGB features) and exceed the performance of all published BEV segmentation models by a margin of 9 points. 

\looseness=-1 This paper has two main contributions: 
First, we elucidate high-impact factors in the design and training protocol of BEV perception models. We show in particular that batch size and input resolution greatly affect performance, while lifting details have a more modest effect. 
Second, we demonstrate that radar data can provide a large boost to performance with a simple fusion method, and invite the community to re-consider this commonly-neglected part of the sensor platform. 
We also release code and reproducible models to facilitate future research in the area. 

\section{Related  Work} \label{sec:related}

A major differentiator in prior work on dense BEV parsing is the precise operator for ``lifting'' 2D perspective-view features to 3D, or directly to the ground plane. 

\looseness=-1 \textbf{Parameter-free unprojection:} This strategy, pursued in a variety of object and scene representation models \cite{activevision,sitzmann2019deepvoxels,commonsense}, uses the camera geometry to define a mapping between voxels and their projected coordinates, and copies 2D features to voxels along their 3D rays. 
We specifically follow the implementation of Harley et al.~\cite{harley2020learning}, which bilinearly samples a subpixel 2D feature for each 3D coordinate. Parameter-free lifting methods are not typically used in bird's eye view parsing tasks. 


\looseness=-1 \textbf{Depth-based unprojection:} Several works estimate per-pixel depth with a monocular depth estimator, either pre-trained for depth estimation \cite{schulter2018learning, liu2020understanding, CaDDN} or trained simply for the end-task \cite{philion2020lift, hu2021fiery, wang2021learning}, and used the depth to place features at their estimated 3D locations. This is an effective strategy, but note that if the depth estimation is perfect, it will only place ``vehicle'' features at the front visible surface of the vehicle, rather than fill the entire vehicle volume with features. We believe this detail is one reason that naive unprojection performs competitively with depth-based unprojection. 

\looseness=-1 \textbf{Homography-based unprojection:} Some works estimate the ground plane instead of per-pixel depth, and use the homography that relates the image to the ground to create a warp \cite{lim2019radar, liu2021weakly, can2022understanding}, transferring the features from one plane to another. This operation tends to produce poor results when the scene itself is non-planar. 

\textbf{MLP-based unprojection:} A popular approach is to convert a vertical-axis strip of image features to a forward-axis strip of ground-plane features, with an MLP \cite{pan2020cross,hendy2020fishing, li2021hdmapnet}. An important detail here is that the initial ground-plane features are considered aligned with the camera frustum, and they are therefore warped into a rectilinear space using the camera intrinsics. Some works in this category use multiple MLPs, dedicated to different scales \cite{roddick2020predicting, saha2021enabling}, or to different categories \cite{gosala2022bird}. As this MLP is parameter-heavy, Yang et al.~\cite{yang2021projecting} propose a cycle-consistency loss (mapping backward to the image-plane features) to help regularize it. 

\looseness=-1 \textbf{Geometry-aware transformer-like models:} An exciting new trend is to transfer features using model components taken from transformer literature. Saha et al.~\cite{saha2021translating} begin by defining a relationship between each vertical scan-line of an image, and the ground-plane line that it projects to, and use self-attention to learn a ``translation'' function between the two coordinate systems. Defining this transformer at the line level provides inductive bias to the model, reusing the lifting method across all lines.  BEVFormer~\cite{li2022bevformer}, which is concurrent work, proposes to use deformable attention operations to collect image features for a pre-defined grid of 3D coordinates. This is similar to the bilinear sampling operation in parameter-free unprojection, but with approximately $10\times$ more samples, learnable offsets for the sampling coordinates, and a learnable kernel on their combination.


\looseness=-1 \textbf{Radar:} In the automotive industry, radar has been in use for several years already~\cite{PARKER2017253}. Since radar measurements provide position, velocity, and angular orientation, the data is typically used to detect obstacles (e.g., for emergency braking), and to estimate the velocity of moving objects (e.g., for cruise control). 
Radar is longer-range and less sensitive to weather effects than LiDAR, and substantially cheaper. Unfortunately, the sparsity and noise inherent to radar make it a challenge to use \cite{lombacher2017semantic,sless2019road,meyer2019deep,sless2019road,yang2020radarnet}. 
Some early methods use radar for BEV semantic segmentation tasks much like in our work~\cite{lombacher2017semantic,schumann2018semantic,sless2019road}, but only in small datasets. Recent work within the nuScenes benchmark~\cite{nuscenes} has reported the data too sparse to be useful, recommending instead higher-density radar data from alternate sensor setups \cite{li2022exploiting, hendy2020fishing}. Some recent works explore RGB-radar or RGB-Lidar fusion strategies \cite{lim2019radar, meyer2019deep}, but focus on detection and velocity estimation rather than BEV semantic labeling.


\begin{figure*}[t]
\centering
 \includegraphics[width=1.0\linewidth]{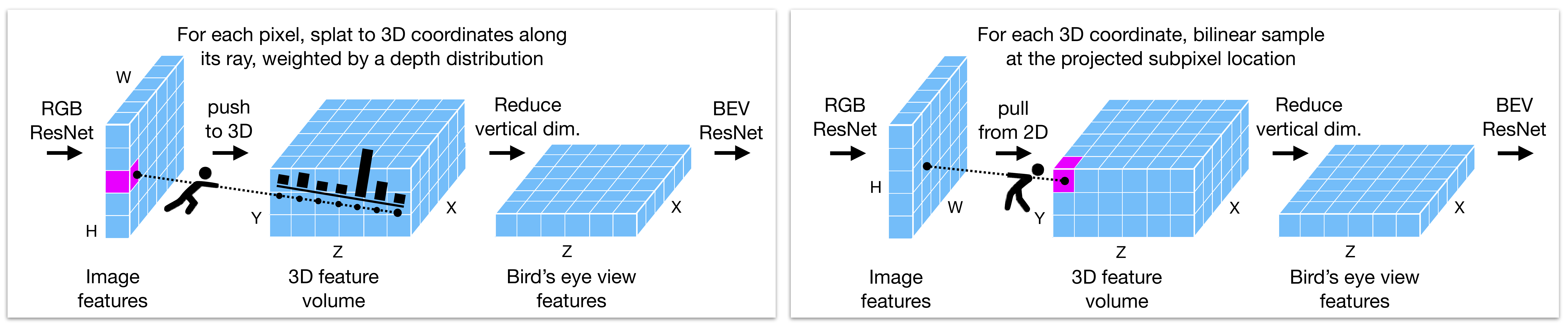}
 \caption{\textbf{2D-to-BEV architecture, illustrated with two lifting strategies.} The left panel shows the Lift-Splat approach \cite{philion2020lift}: in this method, each 2D feature is ``pushed'' to 3D, filling voxels that intersect with its ray. The right panel shows our bilinear sampling approach: in this method, each 3D voxel ``pulls'' a feature from the 2D map, by projection and subpixel sampling.
}
 \label{fig:overview}
\vspace{-1em}
\end{figure*}

\section{Simple-BEV Model}\label{sec:method}

\looseness=-1 In this section, we describe the architecture and training setup of a basic neural BEV mapping model, which we will modify in the experiments to study which factors matter most for performance. 


\subsection{Setup and overview}\label{sec:setup}

\looseness=-1 Our model takes input from cameras, radars, and optionally even LiDAR. 
We assume that the data is synchronized across sensors. We assume that the intrinsics and relative poses of the sensors are known.

\looseness=-1 The model has a 3D metric span, and a 3D resolution. 
Following the baselines in this task, we set the left/right and forward/backward span to $100m \times 100m$, discretized at a resolution of $200 \times 200$. We set the up/down span to $10m$, and discretize at a resolution of $8$. This volume is centered and oriented according to a reference camera, which is typically the front camera. We denote the left-right axis with $X$, the up-down axis with $Y$, and the forward-backward axis with $Z$. 

\looseness=-1 Following related work, we first apply a 2D ResNet to compute features from each camera image, then lift to 3D, then reduce to a BEV plane, and finally apply a 2D ResNet in BEV to arrive at the output. These steps are illustrated in Figure~\ref{fig:overview}, and will be described in more detail in the next section. 
Our lifting step is subtly different from prior work: while some works ``splat'' 2D features along their corresponding 3D rays \cite{philion2020lift,hu2021fiery}, we instead begin at 3D coordinates, and bilinearly sample a sub-pixel feature for each voxel.  
If radar or LiDAR is provided, we rasterize this data into a bird's eye view image, and concatenate this with the 3D feature volume before compressing the vertical dimension. 

\subsection{Architecture design}\label{sec:arch}

We featurize each input RGB image, shaped $3 \times H \times W$, with a ResNet-101~\cite{he2016deep} backbone. We upsample the output of the last layer and concatenate it with the third layer output, and apply two convolution layers with instance normalization and ReLU activations \cite{instancenorm}, arriving at feature maps with shape $C \times H/8 \times W/8$ (one eighth of the image resolution).

We project the pre-defined volume of 3D coordinates into all feature maps, and bilinearly sample there, yielding a 3D feature volume from each camera. We compute a binary ``valid'' volume per camera at the same time, indicating if the 3D coordinate landed within the camera frustum. We then take a valid-weighted average across the set of volumes, reducing our representation down to a single 3D volume of features, shaped $C\times Z\times Y \times X$. 
We then rearrange the axes so that the vertical dimension extends the channel dimension, as in $C\times Z\times Y \times X \rightarrow (C\cdot Y) \times Z\times X$, yielding a high-dimensional BEV feature map. 

\looseness=-1 If radar is provided, we rasterize it to create another BEV feature map with the same spatial dimensions as the RGB-based map. We use an arbitrary number of radar channels $R$ (including $R=0$, meaning no radar). In nuScenes~\cite{nuscenes}, each radar return consists of a total of 18 fields, with 5 of them being position and velocity, and the remainder being the result of built-in pre-processes (e.g., indicating confidence that the return is valid). We use all of this data, by using the position data to choose the nearest $XZ$ position on the grid (if in bounds), and using the 15 non-position items as channels, yielding a BEV feature map shaped $R \times Z \times X$, with $R=15$. If LiDAR is provided, we voxelize it to a binary occupancy grid shaped $Y \times Z \times X$, and use it in place of radar features. 

We then concatenate the RGB features and radar/LiDAR features, and compress the extended channels down to a dimensionality of $C$, by applying a $3 \times 3$ convolution kernel. This achieves the reduction $(C\cdot Y+R)\times Z\times X \rightarrow C\times Z \times X$. 
At this point, we have a single plane of features, representing a bird's eye view of the scene. We process this with three blocks of a ResNet-18~\cite{he2016deep}, producing three feature maps, then use additive skip connections with bilinear upsampling to gradually bring the coarser features to the input resolution, and finally, apply {two convolution layers acting as the segmentation task head.} 
Following FIERY \cite{hu2021fiery}, we complement the segmentation head with auxiliary task heads for predicting centerness and offset, which serve to regularize the model. 
The offset head produces a vector field where, within each object mask, each vector points to the center of that object. We train the segmentation head with a cross-entropy loss, and supervise the centerness and offset fields with an L1 loss. 
We use an uncertainty-based learnable weighting \cite{kendall2018multi} to balance the three losses. 



The 3D resolution is $200 \times 8 \times 200$, and our final output resolution is $200 \times 200$. Our 3D metric span is $100m \times 10m \times 100m$. This corresponds to voxel lengths of $0.5m \times 1.25m \times 0.5m$ (in $Z,Y,X$ order). We use a feature dimension (i.e., channel dimension $C$) of $128$. The ResNet-101 is pre-trained for object detection~\cite{detr} on COCO 2017~\cite{coco2014eccv}. 
The BEV ResNet-18 is trained from scratch. 
We train end-to-end for 25,000 iterations, with the Adam-W optimizer \cite{loshchilov2017decoupled} using a learning rate of 5e-4 and 1-cycle schedule~\cite{smith2019super}.


\subsection{Key factors of study}\label{sec:impl}

\textbf{Lifting strategy:} 
Our model is ``simpler'' than related work, particularly in the 2D-to-3D lifting step, which is handled by (parameter-free) bilinear sampling. This replaces, for example, depth estimation followed by splatting \cite{philion2020lift}, MLPs \cite{pan2020cross, hendy2020fishing, li2021hdmapnet}, or attention mechanisms \cite{saha2021translating,li2022bevformer,zhou2022cross}. 
Our strategy can be understood as ``Lift-Splat \cite{philion2020lift} without depth estimation'', but as as illustrated in Figure~\ref{fig:overview}, our implementation is different in a key detail: our method relies on \textit{sampling} instead of \textit{splatting}. Our method begins with 3D coordinates of the voxels, and takes a bilinear sample for each one. As a result of the camera projection, close-up rows of voxels sample very sparsely from the image (i.e., more spread out), and far-away rows of voxels sample very densely (i.e., more packed together), but each voxel receives a feature.  \textit{Splatting-based} methods \cite{philion2020lift,saha2021translating} begin with a 2D grid of coordinates, and ``shoot'' each pixel along its ray, filling voxels intersected by that ray, at fixed depth intervals. As a result, splatting methods yield multiple samples for up-close voxels, and very few samples (sometimes zero) for far-away voxels. 
As we will show in experiments, this implementation detail has an impact on performance, such that splatting is slightly superior at short distances, and sampling is slightly superior at long distances. In the experiments, we also evaluate a recently-proposed deformable attention strategy~\cite{li2022bevformer}, which is similar to bilinear sampling but with learned sampling kernel for each voxel (i.e., learned weights and learned offsets). 

\textbf{Input resolution:} 
\looseness=-1 While early BEV methods downsampled the RGB substantially before feeding it through the model (e.g., 
downsampling to $128 \times 352$~\cite{philion2020lift}),
we note that recent works have been downsampling less and less (e.g., most recently using the full resolution~\cite{saha2021translating, li2022bevformer}). 
We believe this is an important factor for performance, and so we train and test our RGB-only model across different input resolutions. 
Across variants of our model, we try resolutions from $112 \times 208$ up to $896 \times 1600$. 

\textbf{Batch size:} 
\looseness=-1 Most of the related works on BEV segmentation use relatively small batch sizes in training (e.g., one~\cite{li2022bevformer} or four~\cite{philion2020lift}).
It has been reported in the image classification literature that higher batch sizes deliver superior results~\cite{smith2017don}, but we have not seen batch size discussed as a performance factor in the BEV literature. This may be because of the high memory requirements of these BEV models: it is necessary to process all 6 camera images in parallel with a high-capacity module, and depending on implementation, it is sometimes necessary to store a 3D volume of features before reducing the representation to BEV.
To overcome these memory issues, we accumulate gradients across multiple steps and multiple GPUs, and obtain arbitrarily-large effective batch sizes at the cost of slower wall-clock time per gradient step. For example, it takes us aprrox. 5 seconds to accumulate the forward and backward passes to create a batch size of 40, even using eight A100 GPUs in parallel. Overall, however, our training time is similar to prior work: our model converges in 1-3 days, depending on input resolution. 

\textbf{Augmentations:} 
\looseness=-1 
Prior work recommends the use of camera dropout and various image-based augmentations, but we have not seen these factors quantified. We experiment with multiple augmentations at training time, and measure their independent effects: (1) we apply random resizing and cropping on the RGB input, in a scale range of $[0.8,1.2]$ (and update the intrinsics accordingly), (2) we randomly select a camera to be the ``reference'' camera, which randomizes the orientation of the 3D volume (as well as the orientation of the rasterized annotations), and (3) we randomly drop one of the six cameras. 
At test time, we use the ``front'' camera as the reference camera, and do not crop. 

\textbf{Radar usage details:} 
Prior work has reported that the radar data in nuScenes is too sparse to be useful~\cite{li2022exploiting, hendy2020fishing}, but we hypothesize it can be a valuable source of metric information, lacking from current camera-only setups. Beyond simply using radar or not, our model is flexible in terms of (1) using the radar meta-data as additional channels, vs. treating the radar as a binary occupancy image, (2) using the raw data from the sensor, vs. using outlier-filtered input, and (3) using radar accumulated from a number of sweeps, vs. only using the time-synchronized data. 

\section{Experiments}\label{sec:exp}

Our experiments first aim to provide a unified study of the factors affecting BEV segmentation model performance, including high-interest details such as lifting strategies, and rarely-discussed details such as resolution and batch size. Second, we aim to quantify the utility of radar in this domain, and reveal the usage details which maximize performance. Finally, we compare against the state-of-the-art. 



\looseness=-1 \textbf{Dataset:} We train and test all models in the nuScenes~\cite{nuscenes}
urban scenes {dataset}, which {is} publicly available for non-commercial use. {The dataset has} 6 cameras, pointing front, front-left, front-right, back-left, back, and back-right, and 5 radar units, pointing front, left, right, back-left, and back-right, as well as a LiDAR unit. We use LiDAR inputs only for comparison, and focus on using RGB and RGB+radar. 
We use the official nuScenes training/validation split, which contains 28,130 samples in the training set, and 6,019 samples in the validation set. 
We follow the segmentation task setup from the official Lift-Splat~\cite{philion2020lift} codebase, where any point within a ``vehicle'' bounding box is counted as a positive label, and all other locations are given negative labels. The ``vehicle'' superclass  consists of bicycle, bus, car, construction vehicle, emergency vehicle, motorcycle, trailer, and truck. 
We evaluate using the intersection-over-union (IOU) metric. 

\subsection{Unified study of performance factors}

\begin{figure}[t]
\centering
 \includegraphics[width=0.7\linewidth]{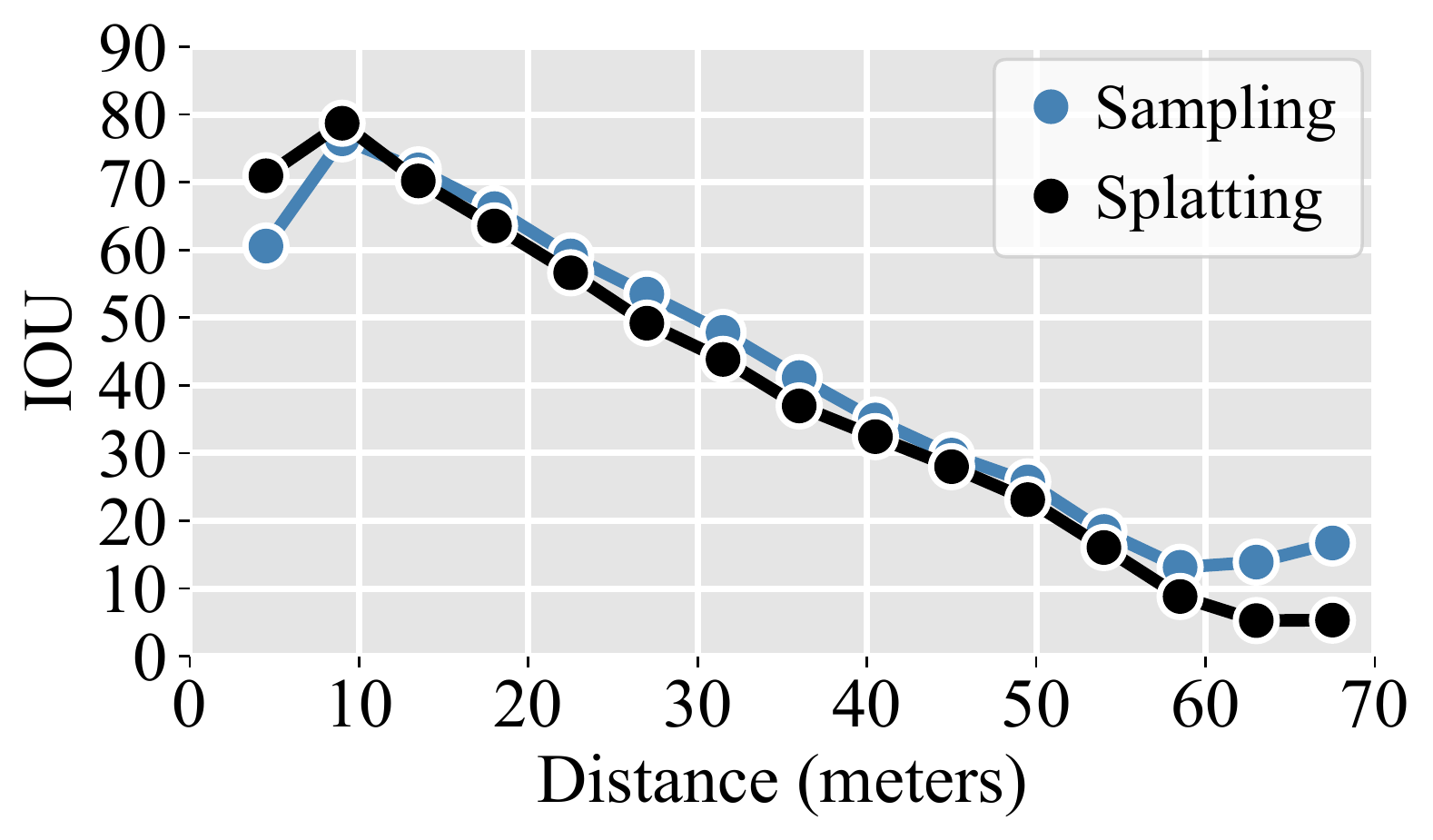}
 \caption{Comparing IOU over distance for Lift-Splat-style splatting versus our bilinear sampling strategy, splatting is better at close range, while bilinear sampling is better at medium to long range.}
 \label{fig:iou_over_dist}
\vspace{-1em}
\end{figure}

\begin{table}[b]
\centering
\caption{Effect of lifting strategy.}\label{tab:lifting}
\begin{tabular}{lc}
\toprule
2D-to-BEV strategy & IOU \\
\hline
{Unweighted splatting} & {43.1} \\
{Depth-based splatting \cite{philion2020lift}} & {44.4} \\
{Deformable attention} \cite{li2022bevformer} & {46.5} \\
{Bilinear sampling (ours)} & \underline{47.4} \\
{Multi-scale deform. attn. \cite{li2022bevformer}} & {\textbf{48.9}} \\
\bottomrule
\end{tabular}
\end{table}
\looseness=-1 \textbf{Lifting strategy:} Our model uses a bilinear sampling strategy to lift image features onto the BEV plane. 
Prior work has developed a variety of sophisticated alternate methods, but since various implementation details change across each work, it is unclear how much the lifting methods differ in performance. We present an apples-to-apples comparison in Table~\ref{tab:lifting}, matching resolution, batch size, backbone, and augmentations across models. 
We see that bilinear sampling and deformable attention perform similarly, while the splatting methods fall behind. 
Multi-scale deformable attention (as in BEVFormer \cite{li2022bevformer}) performs best, but at the cost of speed (1 day slower to train, 0.5 FPS slower at test time) and complexity (59M parameters instead of 42M, and requiring a custom CUDA kernel). 
We also note that uniform (unweighted) splatting is only approximately 1 point worse than depth-weighted splatting (consistent with results observed in FIERY~\cite{hu2021fiery}), suggesting that much of the scene structure is resolved after approximate BEV lifting. 
Fig.~\ref{fig:iou_over_dist} shows a breakdown of IOU across distance for sampling vs. depth-weighted splatting, revealing that splatting provides an advantage at short distances, while sampling is better at long distances. 
We note that all results in our evaluation are higher than reported in the works introducing these methods, suggesting that other details in our training setup are having a substantial impact. As we show in the next subsections, the major factors here are input resolution and batch size. 

\begin{figure}[t]
\begin{center}
\includegraphics[width=0.65\linewidth]{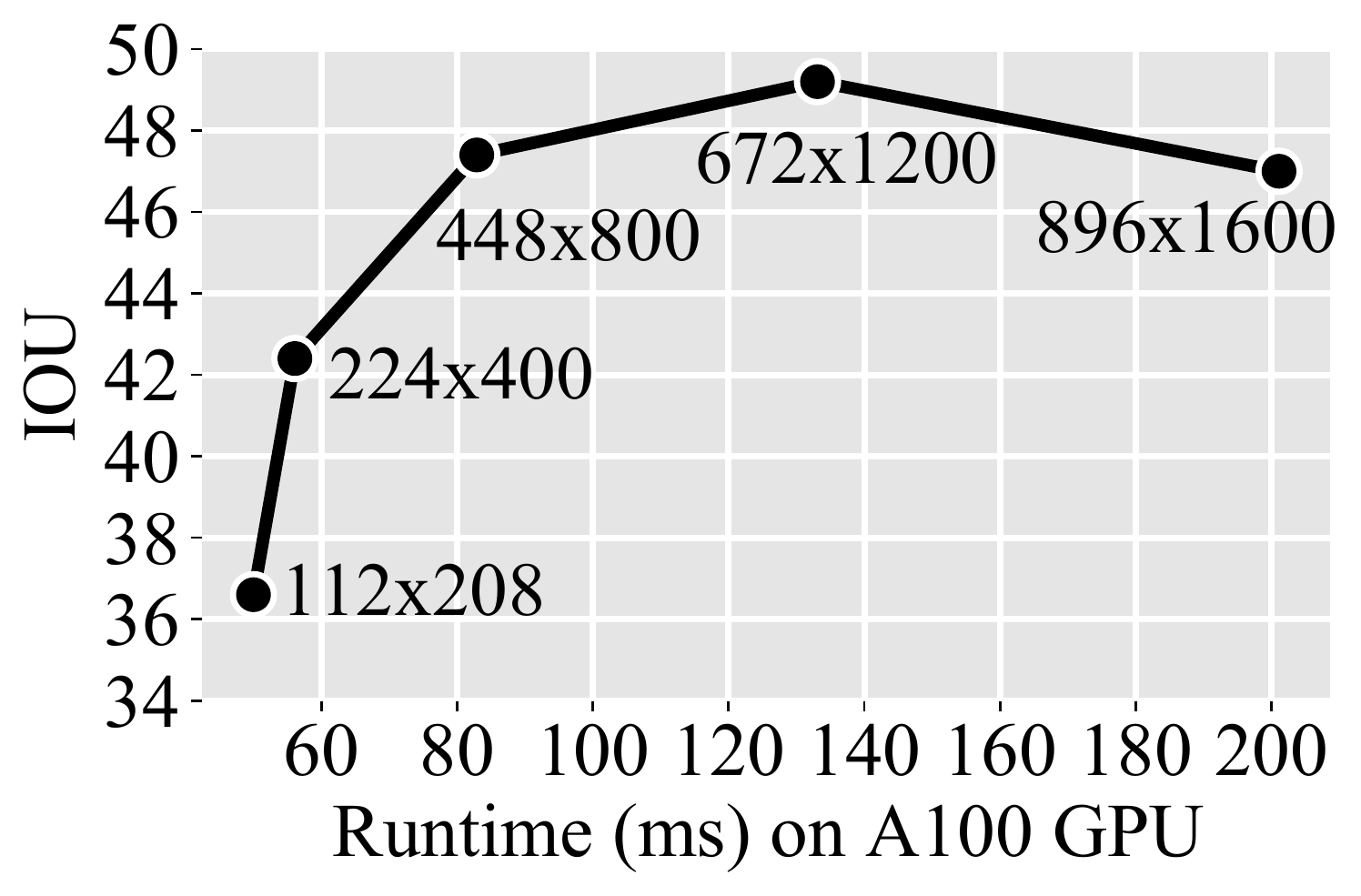}
\end{center}
\vspace{-1em}
\caption{Effect of input resolution.}
\vspace{-1em}
\label{fig:resolutions}
\end{figure}

\looseness=-1 \textbf{Input resolution:} 
We measure how our model's performance changes with input resolution, using the same resolution to train and test. 
Fig.~\ref{fig:resolutions} summarizes the results. {Using resolutions lower than $448 \times 800$ drastically worsens performance.}
Our best result is \textbf{49.3} IOU at $672 \times 1200$. However, this model is substantially slower than the 47.4 IOU $448 \times 800$ model (133 ms vs 83 ms), and requires nearly twice the training time.
Results drop at the highest resolution, perhaps because when the images are this large, the object scale is no longer consistent with the backbone's pre-training, leading to less-effective transfer. 
{Performance at high resolution may improve with an alternate backbone architecture~\cite{park2021pseudo}.}



\begin{figure}[b]
\begin{center}
    \includegraphics[width=0.65\linewidth]{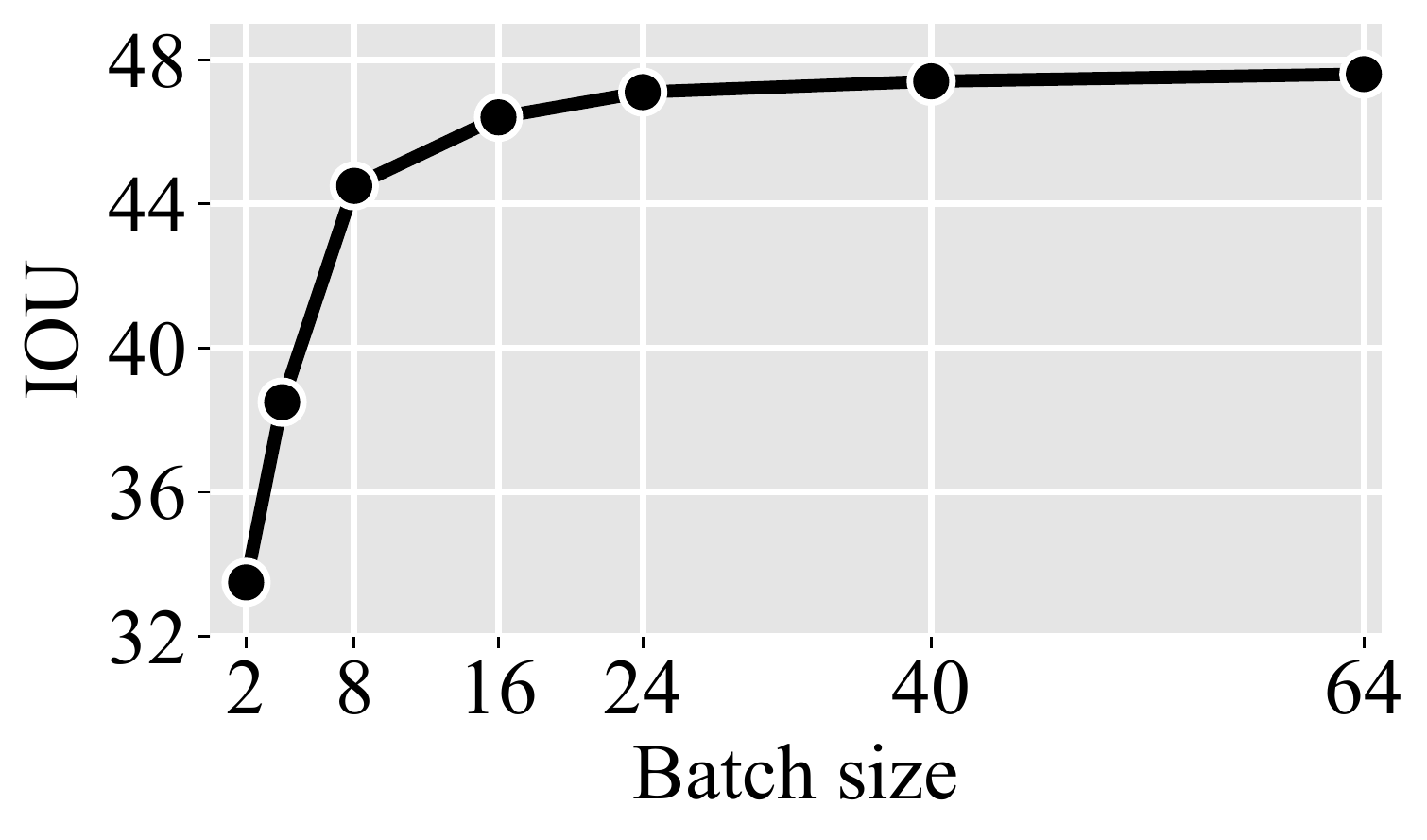}
\end{center}
\vspace{-1em}
\caption{Effect of batch size.}
\label{fig:batch}
\end{figure}

\textbf{Batch size:} 
In Fig.~\ref{fig:batch} we explore the impact of batch size on our model's performance:
each increase in batch size gives an improvement in accuracy, with diminishing (but sizeable) returns. Increasing the batch size from 2 to 40 gives a nearly 14-point improvement in IOU. 
Considering that most prior works used batch sizes under 16, this suggests that many existing methods may benefit from simply re-training.

{
\textbf{Backbones:} Recent works have been using deeper and deeper backbones for the part of the model that creates feature maps from the input camera images (before the 2D-to-BEV lifting). Lift-Splat~\cite{philion2020lift} used an EfficientNet-B0~\cite{tan2019efficientnet}, FIERY~\cite{hu2021fiery} used an EfficientNet-B4~\cite{tan2019efficientnet}, TIIM~\cite{saha2021translating} used a ResNet-50~\cite{he2016deep}, and recently BEVFormer~\cite{li2022bevformer} used a ResNet-101~\cite{he2016deep}. We explore these choices for our own model in Table~\ref{tab:backbones}. We find, as expected, that a larger backbone gives better results, with ResNet-101 outperforming the rest. 
However, we note that the benefit of a specific backbone is sometimes tied to the input resolution, which is fixed here at $448 \times 800$. Making the best use of the available $900 \times 1600$ data may require further exploration into architecture details.
}

\begin{table}[t]
\centering
\caption{{Effect of backbone.}}
\begin{tabular}{lc}
\toprule
{Backbone} & {IOU} \\
\hline
EfficientNet-B0 & 43.7 \\
EfficientNet-B4 & 46.4 \\
ResNet-50 & \underline{46.6} \\
ResNet-101 & \textbf{47.4} \\
\bottomrule
\end{tabular}
\label{tab:backbones}
\vspace{-1em}
\end{table}

\begin{table}[b]
\caption{Study of augmentations.}
    \begin{subtable}[h]{0.3\linewidth}
    \centering
    \begin{tabular}{lc}
    \toprule
    {Crop/resize} & {IOU} \\
    \midrule
    Off & 45.8 \\
    On & \textbf{47.4} \\
    \bottomrule
    \end{tabular}
    \vspace{0.5em}  
    \caption{Effect of augmentations.}\label{tab:augs}
    \end{subtable}
\hfill
    \begin{subtable}[h]{0.32\linewidth}
    \centering
    \begin{tabular}{lc}
    \toprule
    {Ref. camera} & {IOU} \\
    \midrule
    ``Front'' & 46.8 \\
    Random & \textbf{47.4} \\
    \bottomrule
    \end{tabular}
    \vspace{0.5em}  
    \caption{Effect of camera randomization. }\label{tab:refcam}
    \end{subtable}
\hfill
    \begin{subtable}[h]{0.3\linewidth}
    \centering
    \begin{tabular}{lc}
    \toprule
    {\# Cameras} & {IOU} \\
    \midrule
    5/6 & 46.4 \\
    6/6 & \textbf{47.4} \\
    \bottomrule
    \end{tabular}
    \vspace{0.5em}  
    \caption{Effect of camera dropout.}\label{tab:ncams}
    \end{subtable}
\end{table}

\textbf{Augmentations:} 
When training our model, we randomly resize each camera's image to within $[0.8,1.2]$ of the target resolution, and place it at a random offset from the center. Table~\ref{tab:augs} shows that this augmentation gives a 1.6 point boost in IOU.
When also randomize the camera selected to be the ``reference'' camera, which dictates the orientation of the 3D coordinate system. 
We show the results of this augmentation in Table~\ref{tab:refcam}. Randomizing the reference camera provides a 0.6 point boost in IOU. We believe that randomizing the reference camera helps reduce overfitting in the bird's eye view module. 
We have observed qualitatively that without this augmentation, the segmented cars have a slight bias for certain orientations in certain positions; with the augmentation added, this bias disappears. Prior work has reported a benefit from randomly dropping 1 of the 6 available cameras in each training sample~\cite{philion2020lift}. As shown in Table~\ref{tab:ncams}, we find the opposite: using all cameras performs 1 point better. It may be that our reference-camera randomization provides enough regularization to make camera-dropout unnecessary. 
We have also experimented with photometric augmentations (blur, color, contrast), 
but found that they did not help. 

\subsection{Multi-modality fusion analysis}
\begin{figure}[t]
\centering
 \includegraphics[width=1.0\linewidth]{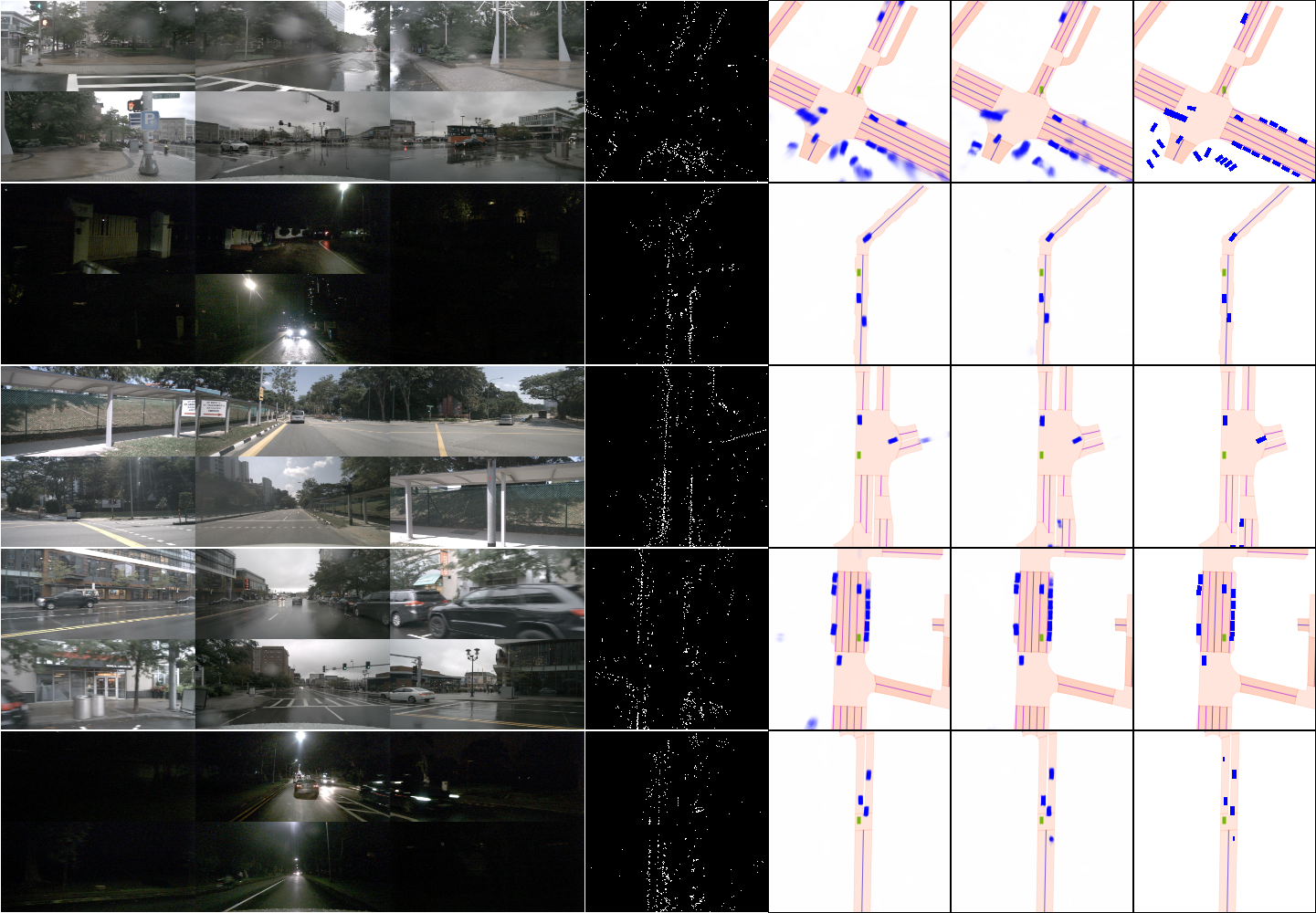}
 \caption{
 {Left to right: camera data, radar data, our RGB model output, our RGB+radar model output, and ground truth.} 
 }
 \label{fig:results}
\vspace{-1em}
\end{figure}

To analyze performance across modality combinations, we compare camera-only vs. camera plus radar vs. camera plus LiDAR, in Table~\ref{tab:modalities}. As hypothesized, radar indeed improves results in our setup. 
Relative to the camera-only model, radar improves results by 8 points, and LiDAR improves results by 13 points. High performance from LiDAR is consistent with related work in 3D object detection~\cite{yin2021center}, but the gap between RGB+LiDAR and RGB+radar is smaller than might have been expected, since prior work conveyed negative results from RGB+radar fusion~\cite{hendy2020fishing}. 
We show qualitative results in Fig.~\ref{fig:results}. We also visualize corresponding radar data. 
Qualitatively, the radar is indeed sparse and noisy as noted in related work \cite{hendy2020fishing, li2022exploiting}, but we believe it gives valuable hints about the metric scene structure, which, when fused with information acquired from RGB, enables higher-accuracy semantic segmentation in the bird's eye view. 
We next investigate the radar performance factors in more detail. 


\begin{table}[t]
\centering
\caption{{Effect of multi-modality fusion.}}\label{tab:modalities}
\begin{tabular}{lc}
\toprule
{Modality} & {IOU} \\
\hline
Camera & 47.4 \\
Camera + radar & \underline{55.7} \\
Camera + LiDAR & \textbf{60.8} \\
\bottomrule
\end{tabular}
\vspace{-1em}
\end{table}
\looseness=-1 As shown in Table~\ref{tab:radar_meta}, our model benefits from accessing the meta-data associated with each radar point. This includes information such as velocity, which may help distinguish moving objects from the background. Removing this aspect lowers IOU by 0.7 points. As shown in Table~\ref{tab:radar_filter}, our model benefits from having \textit{all} radar returns as input, achieved by disabling nuScenes' built-in outlier filtering strategy. The filtering strategy attempts to discard outlier points (produced by multipath interference and other issues), but potentially discards some true returns as well. 
Using the filtered data instead of the raw data results in a 2 point drop in performance. As shown in Table~\ref{tab:radar_sweeps}, our model benefits from aggregating multiple sweeps of radar as input. This means using radar from timesteps $(t, t-1, t-2)$ aligned to the coordinate frame of timestep $t$, rather than exclusively using the data from timestep $t$. Using a single sweep lowers performance by 2.4 points, likely because the model struggles with the extreme sparsity of the signal.

\begin{table}[b]
\caption{Study of radar hyperparameters.}
    \begin{subtable}[h]{0.30\linewidth}
    \centering
    \begin{tabular}{cc}
    \toprule
    {Input} & {IOU} \\
    \midrule
    Occ. only & {55.0} \\
    Full return & \textbf{55.7} \\
    \bottomrule
    \end{tabular}
    \vspace{0.5em}
    \caption{Using meta-data associated with points helps.}\label{tab:radar_meta}
    \end{subtable}
\hfill
    \begin{subtable}[h]{0.30\linewidth}
    \centering
    \begin{tabular}{cc}
    \toprule
    {Filtering} & {IOU} \\
    \midrule
    On & {53.7} \\
    Off & \textbf{55.7} \\
    \bottomrule
    \end{tabular}
    \vspace{0.5em}
    \caption{Disabling nuScenes' outlier-filtering helps.}\label{tab:radar_filter}
    \end{subtable}
\hfill
    \begin{subtable}[h]{0.3\linewidth}
    \centering
    \begin{tabular}{cc}
    \toprule
    {\# Sweeps} & {IOU} \\
    \midrule
    1 & {53.1} \\
    3 & \textbf{55.7} \\
    \bottomrule
    \end{tabular}
    \vspace{0.5em}
    \caption{Aggregating multiple radar sweeps helps.}\label{tab:radar_sweeps}
    \end{subtable}
\label{tab:radar}
\end{table}



\subsection{Comparison with state-of-the-art}



\begin{table}[t]
\setlength{\tabcolsep}{2.5pt}
\centering
\caption{Comparison against {state-of-the-art} methods for vehicle segmentation IOU in the nuScenes validation set.}\label{tab:sota}
\begin{tabular}{lccccc}
\toprule
{Method} & Lifting & Batch & Resolution & {Inputs} & {IOU} \\
\midrule
\multirow{2}{*}{FISHING \cite{hendy2020fishing}} & \multirow{2}{*}{MLP} & \multirow{2}{*}{-} & \multirow{2}{*}{$192 \times 320$} & RGB & 30.0 \\
 &&&& \gray{LiDAR} & \gray{(44.3)} \\
\hline
\multirow{2}{*}{Lift, Splat \cite{philion2020lift}} & \multirow{2}{*}{Depth splat} &\multirow{2}{*}{4} &\multirow{2}{*}{$128 \times 352$} & RGB & 32.1 \\
&&&& \gray{RGB+LiDAR} & \gray{(44.5)} \\
\hline
\multirow{2}{*}{FIERY \cite{hu2021fiery}} & \multirow{2}{*}{Depth splat} &\multirow{2}{*}{12} &\multirow{2}{*}{$224\times 480$} & RGB & 35.8 \\
 &&&& RGB+time & 38.2 \\
\hline
\multirow{2}{*}{TIIM \cite{saha2021translating}} & \multirow{2}{*}{Ray attn.} &\multirow{2}{*}{8} &\multirow{2}{*}{$900 \times 1600$} & RGB & 38.9\\
 &&&& RGB+time & 41.3\\
\hline 
{CVT \cite{zhou2022cross}} & {Attention} &{16} &{$224 \times 448$} & RGB & 36.0\\
\hline
\multirow{2}{*}{BEVFormer \cite{li2022bevformer}} & \multirow{2}{*}{Def. Attn.} &\multirow{2}{*}{1} &\multirow{2}{*}{$900 \times 1600$} & RGB & 44.4\\
 &&&& RGB+time & 46.7\\
\hline
\multirow{3}{*}{Ours} & \multirow{3}{*}{Bilinear} & \multirow{3}{*}{40} &\multirow{3}{*}{$448\times800$} & RGB & \underline{47.4}\\
 &&&& RGB+radar & \textbf{55.7}\\
 &&&& \gray{RGB+LiDAR} & \gray{(\textbf{60.8})} \\
\bottomrule
\end{tabular}
\end{table}

\looseness=-1 In this final subsection, we compare against the published state-of-the-art. Considering that our previous experiments show the importance of various subtle training details (which vary across each prior work), the comparisons in this section should not be read as an apples-to-apples comparison, but rather  as reflecting a combination of factors in the training setups. Besides the factors already considered, we note that some models train for additional categories, which may either improve \textit{or} worsen results \cite{li2022bevformer}. 
In Table~\ref{tab:sota}, we see that our RGB model slightly outperforms all other RGB-based and temporal models, while the RGB+radar model holds a margin of 9 points above the rest. 
Our model has 42M parameters, which is fairly efficient compared to BEVFormer's 68.7M. Most of our parameters (37M) come from the ResNet-101, and this is also the main speed bottleneck. 
Our model is also 3 times faster than BEVFormer in this setup (7.3 FPS vs 2.3 FPS on a V100 GPU), largely due to using a lower RGB resolution.
 %


\section{Discussion and conclusion}
\looseness=-1 In this work, we explore design and training choices for BEV semantic parsing, and show that batch size and image resolution play a surprisingly large role in performance, which has not been previously discussed in the literature. While recent work has developed increasingly sophisticated 2D-to-BEV lifting strategies, we show that bilinear sampling performs well also. 
Temporal integration is a natural fit in this setting, but we leave it for future work. 
Making best use of the available high-resolution images will require a more thorough exploration of backbones. 
Another area for future work is to explore 3D object detection, in addition to (or instead of) the dense BEV representation.
Our experiments demonstrate that radar provides useful information for BEV parsing, and we hope that this insight will be applied to other approaches. 
Our work does not argue for the use of particular sensors over others, but rather for the use of metric information whenever available, even if sparse and noisy.

\bibliographystyle{IEEEtran}
\bibliography{99_refs}

\end{document}